%% file: main.tex
\documentclass[runningheads]{llncs}

\usepackage[mobile]{eccv}

\usepackage{eccvabbrv}
\usepackage{adjustbox}
\usepackage{graphicx}
\usepackage{booktabs}
\usepackage{makecell}
\usepackage{xcolor}
\usepackage{amssymb}
\usepackage{pifont}
\usepackage{wrapfig}
\usepackage{caption}
\usepackage{subcaption}
\usepackage[accsupp]{axessibility}  %

\usepackage[many]{tcolorbox}

\definecolor{mybgcolor}{HTML}{F6F7F2}
\definecolor{myframecolor}{HTML}{A5C2A5}
\definecolor{myshadowcolor}{HTML}{E2E6D9}

\newtcolorbox{highlightbox}{
    enhanced,
    colback=mybgcolor,
    colframe=myframecolor,
    boxrule=1pt,
    arc=3mm,
    drop shadow=myshadowcolor,
    left=10pt,
    right=10pt,
    top=5pt,
    bottom=5pt,
    fontupper=\itshape
}

\definecolor{brownbgcolor}{HTML}{FBF7F3}
\definecolor{brownframecolor}{HTML}{B89A82}
\definecolor{brownshadowcolor}{HTML}{E6DCD3}

\newtcolorbox{promptbox}{
    enhanced,
    title=Prompts,
    fonttitle=\bfseries,
    colback=brownbgcolor,
    colframe=brownframecolor,
    boxrule=1pt,
    arc=3mm,
    drop shadow=brownshadowcolor,
    left=10pt,
    right=10pt,
    top=5pt,
    bottom=5pt,
    fontupper=\itshape
}

\newcommand{\cmark}{\ding{51}}
\newcommand{\xmark}{\ding{55}}

\usepackage[pagebackref=true,breaklinks=true,colorlinks,bookmarks=false,allcolors=eccvblue]{hyperref}
\usepackage{fontawesome5}
\usepackage{multirow}
\usepackage{colortbl}
\usepackage[capitalize,noabbrev]{cleveref}
\crefname{figure}{Fig.}{Figs.}
\Crefname{figure}{Fig.}{Figs.}
\crefname{table}{Tab.}{Tabs.}
\Crefname{table}{Tab.}{Tabs.}
\crefname{section}{Sec.}{Secs.}
\Crefname{section}{Sec.}{Secs.}
\usepackage{orcidlink}

\definecolor{lyellow}{HTML}{F7F2CC}
\definecolor{lpink}{HTML}{F6D6D6}
\definecolor{mycol1}{HTML}{F2EEEC}
\definecolor{mycol2}{HTML}{7E6A60}

\newcommand{\smallrocket}{\raisebox{0.15ex}{\scalebox{0.75}{\faRocket}}}

\begin{document}

\title{\emph{LVOmniBench}: Pioneering Long Audio-Video Understanding Evaluation for Omnimodal LLMs} 

\titlerunning{Abbreviated paper title}

\author{
    Keda Tao\textsuperscript{1,2,3,$\dagger$}\;
    Yuhua Zheng\textsuperscript{1,2,$\dagger,\ddagger$}\;
    Jia Xu\textsuperscript{2,$\ddagger$}\; \\
    Wenjie Du\textsuperscript{2,$\ddagger$}\;
    Kele Shao\textsuperscript{1,4,2}\;
    Hesong Wang\textsuperscript{1,2}\;
    Xueyi Chen\textsuperscript{2}\;
    Xin Jin\textsuperscript{2}\; \\
    Junhan Zhu\textsuperscript{2,$\ddagger$}\;
    Bohan Yu\textsuperscript{3}\;
    Weiqiang Wang\textsuperscript{3}\;
    Jian Liu\textsuperscript{3}\;
    Can Qin\textsuperscript{5}\; \\
    Yulun Zhang\textsuperscript{6}\;
    Ming-Hsuan Yang\textsuperscript{7}\;
    Huan Wang\textsuperscript{2,$\star$} \\
    \vspace{1mm}
    \textsuperscript{1} Zhejiang University\;
    \textsuperscript{2} Westlake University\;
    \textsuperscript{3} Ant Group\; \\
    \textsuperscript{4} Shanghai Innovation Institute\;
    \textsuperscript{5} Independent Researcher\;\\
    \textsuperscript{6} Shanghai Jiao Tong University\;
    \textsuperscript{7} University of California Merced \\
}

\institute{\textcolor{magenta}{\tt{ \url{https://kd-tao.github.io/LVOmniBench/}}}}

\maketitle

\vspace{-5mm}
\input{sec/0_abs}
\input{sec/1_intro}
\input{sec/2_related}
\input{sec/3_benchmark}

\input{sec/4_exp}
\input{sec/5_conclusion}

\bibliographystyle{splncs04}
\bibliography{main}

\input{sec/supp}

\end{document}

%% file: sec/0_abs.tex
\begin{abstract}
Recent advancements in omnimodal large language models (OmniLLMs) have significantly improved the comprehension of audio and video inputs.
However, current evaluations primarily focus on short audio and video clips ranging from 10 seconds to 5 minutes, failing to reflect the demands of real-world applications, where videos typically run for tens of minutes.
To address this critical gap, we introduce \textbf{LVOmniBench}, a new benchmark designed specifically for the cross-modal comprehension of \textit{long-form} audio and video.
This dataset comprises high-quality videos sourced from open platforms that feature rich audio-visual dynamics.
Through rigorous manual selection and annotation, LVOmniBench comprises 275 videos, ranging in duration from 10 to 90 minutes, and 1,014 question-answer (QA) pairs.
LVOmniBench aims to rigorously evaluate the capabilities of OmniLLMs across domains, including long-term memory, temporal localization, fine-grained understanding, and multimodal perception.
Our extensive evaluation reveals that current OmniLLMs encounter significant challenges when processing extended audio-visual inputs.
Open-source models generally achieve accuracies below 35\%, whereas the Gemini 3 Pro reaches a peak accuracy of approximately 65\%.
We anticipate that this dataset, along with our empirical findings, will stimulate further research and the development of advanced models capable of resolving complex cross-modal understanding problems within long-form audio-visual contexts.

\keywords{OmniLLMs \and Audio-Video Understanding \and Benchmark}

\renewcommand{\thefootnote}{\fnsymbol{footnote}}
\vspace{1mm}
\makeatletter
\renewcommand{\@makefntext}[1]{%
  \parindent 1em\noindent
  \hbox to 1.8em{#1\hss}} %

\makeatother
\footnotetext[1]{
$^\star$Corresponding author: Huan Wang (\texttt{wanghuan@westlake.edu.cn}).}
\footnotetext[2]{$^\dagger$ Equal Contribution.}
\footnotetext[3]{$^\ddagger$ Work done as a visiting research intern at ENCODE Lab, Westlake University.
}

\end{abstract}

%% file: sec/1_intro.tex
\section{Introduction}
\label{sec:intro}

The rapid development of omnimodal large language models (OmniLLMs) has highlighted their significant perceptual and cognitive capabilities in integrating vision, audio, and text~\cite{xu2025qwen2,xu2025qwen3,sun2024video,tang2025video,damonlpsg2024videollama2,fu2025vita,ye2025omnivinci,li2024baichuanomni,chen2025chronusomni,yang2025humanomniv2,tong2025interactiveomni,team2025longcat,ai2025ming,yao2024minicpm,fu2024vita}.
These advancements demonstrate the substantial potential of OmniLLMs as foundation models that can simultaneously comprehend real-world audio and video inputs.
However, real-world videos are not merely combinations of isolated modalities; they are intrinsically long-form, often spanning tens of minutes and featuring highly intertwined audio-visual streams.
This extended temporal dimension amplifies the complexity of multimodal interactions, posing significant challenges for fine-grained understanding, cross-modal alignment, and reasoning.
Consequently, while current OmniLLMs perform well on isolated tasks, they struggle with highly complex scenarios.
Furthermore, although numerous benchmarks exist for omnimodal understanding~\cite{chen2025uno,li2024omnibench,zhou2025daily,hong2025worldsense,li2025omnivideobench,sung2024avhbench,chao2025jointavbench,cao2025xgc,wang2025lvbench,zhang2025omnieval,ma2025fortisavqa}, most are limited to reasoning over static image-audio pairs~\cite{li2024omnibench,gong2024av} or short-form video clips~\cite{zhou2025daily,hong2025worldsense,li2025omnivideobench,sung2024avhbench,chao2025jointavbench}.
This scarcity of evaluations for long-form joint audio-video content leaves a critical gap in comprehensively assessing and advancing robust OmniLLMs for real-world applications.

\begin{figure}[t]
    \centering
    \includegraphics[width=\linewidth]{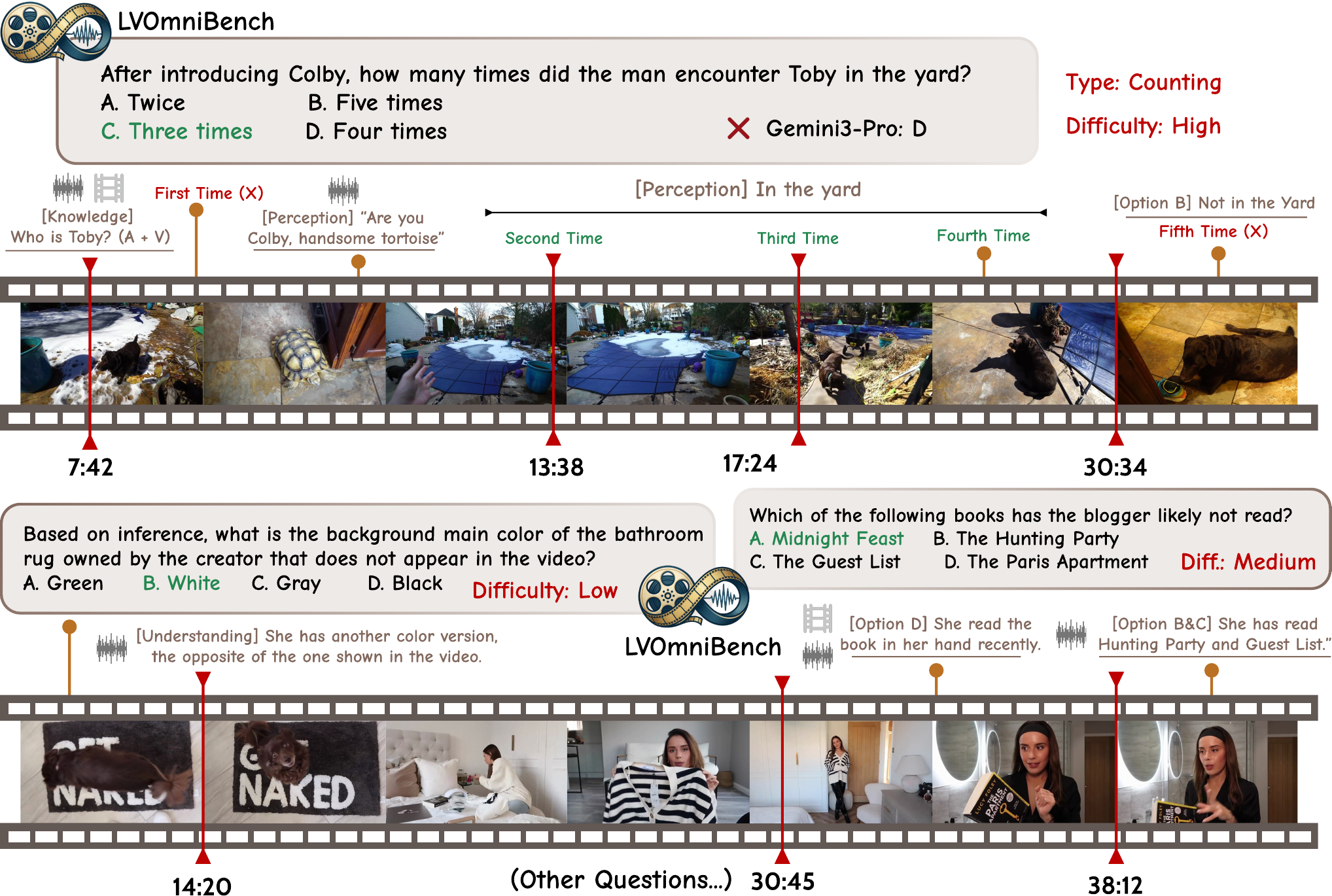}
    \vspace{-6mm}
    \caption{
We introduce \textcolor{mycol2!80!black}{\textbf{LVOmniBench}} to provide a rigorous assessment of the performance of OmniLLMs on \textit{long} audio-visual inputs, comprising strictly manually curated videos and annotations.
Each question is assigned a specific difficulty level to facilitate hierarchical evaluation of model performance.
In the first example, the model should comprehend the entire audio-visual context.
Solving this question initially requires cross-modal alignment to identify ``Toby'' and finally necessitates visual counting and scene recognition.
Even SoTA models, such as Gemini 3 Pro, struggle to answer this question correctly.
The second example presents questions at two additional difficulty tiers, demonstrating that a correct answer requires a combination of audio and visual.
}
\label{fig:benchmark_case_teaser}
\vspace{-7mm}
\end{figure}

To this end, we introduce \textbf{LVOmniBench}, which, to the best of our knowledge, is the first comprehensive benchmark specifically designed for the rigorous evaluation of OmniLLMs in understanding long-form, integrated audio-visual content.
We curated a dataset comprising 275 long videos across diverse scenarios, totaling 140 hours; the scale of this dataset significantly surpasses that of previous audio-visual benchmarks while ensuring rich spatiotemporal and acoustic dynamics within each video.
In addition, to ensure broad generalization, the dataset encompasses a wide range of categories.
We manually constructed 1,014 high-quality multiple-choice questions, which are explicitly designed to require joint reasoning across the audio and visual modalities, thereby facilitating a more comprehensive evaluation of OmniLLMs.
\cref{fig:benchmark_case_teaser} illustrates three representative question paradigms within the benchmark, which are categorized across varying levels of difficulty.
As demonstrated, each task strictly necessitates cross-modal audio-visual reasoning.
Furthermore, these examples highlight the inherent challenges of OmniLLMs when processing extended audio-visual inputs.

\input{tab/benchmark_compare}

Based on extensive experiments and the rigorously constructed benchmark, the principal contributions and findings are summarized as follows:
\begin{itemize}
\item We introduce LVOmniBench, a new benchmark specifically designed to evaluate OmniLLMs for cross-modal comprehension of long-form audio-visual content, constructed through strictly manual video curation and annotation.
\item We curated a diverse collection of long videos, with durations ranging from 10 to 90 minutes and an average duration of 2,069s.
This duration represents a greater than sixfold increase in temporal scale compared to that of existing benchmarks for audio-visual understanding (as illustrated in \cref{tab:benchmark_compare}).
\item Within LVOmniBench, each question is classified into multiple types, including perception, understanding, inference, and complex logical reasoning, and is assigned one of three levels of difficulty, thereby allowing for a hierarchical evaluation of the performance of models.
\item Experimental results demonstrate that the processing of long audio-visual sequences remains a significant challenge for OmniLLMs.
Even the SoTA model, Gemini 3 Pro, achieves a peak accuracy of only 65\%, whereas open-source counterparts struggle to surpass 35\%, yielding a performance that is often marginally better than random chance.
These findings underscore the necessity for further improvements in the processing of extended audio-visual inputs and cross-modal alignment.
\end{itemize}

%% file: tab/benchmark_compare.tex
\begin{table*}[t]
\centering
\caption{Comparison of different benchmarks. The modalities V, I, and A denote video, image, and audio, respectively. The open-domain metric indicates whether the video sources span a broad range of real-world scenarios. The annotation method (Anno.) specifies whether the dataset was curated manually (M) or automatically (A). We also report the video duration and the average temporal duration for each dataset, both measured in seconds. Multi-dimension (Multi-D) indicates whether the benchmark questions are hierarchically stratified by varying levels of cognitive difficulty, and multi-type indicates whether the problem involves multiple aspects.}
\vspace{-2mm}
\setlength{\tabcolsep}{3pt}
\resizebox{\textwidth}{!}{
\begin{tabular}{l c c c c c c c c c}
\toprule
\textbf{Benchmark} & \textbf{Modality} & \makecell{\textbf{Open}\\\textbf{Domain}} &  \makecell{\textbf{Video}\\\textbf{Duration}} & \textbf{Avg. Length} & \textbf{Anno.} & \textbf{Multi-Type} & \textbf{Multi-D} &  \makecell{\textbf{Answer}\\\textbf{Type}} \\
\midrule
AVQA~\cite{yang2022avqa} & V+A & \xmark & 10 & 10 & M & \xmark & \xmark & MC \\
Music-AVQA~\cite{li2022learning} & V+A & \xmark & 60 & 60 & M & \xmark & \xmark & CLS \\
AVTrustBench~\cite{chowdhury2025avtrustbench} & V+A & \cmark & 10\textbackslash 60 & - & A\&M & \xmark & \xmark & MC \\
AV-Odyssey~\cite{gong2024av} & I+A & \cmark & - & - & M & \cmark & \xmark & MC \\
AVHBench~\cite{sung2024avhbench}  & V+A & \cmark & 10 & 10 & A\&M & \cmark & \xmark & CLS \\
OmniBench~\cite{li2024omnibench} & I+A & \cmark & - & - & M & \cmark & \xmark & MC \\
LongVALE~\cite{geng2025longvale} & A+V & \cmark & 30--600 
& 235 & A\&M & \xmark & \xmark & OE\\
AVUT~\cite{yang2025audio} & V+A & \cmark & - & 67.8 & A\&M & \cmark & \xmark & MC\\
Daily-Omni~\cite{zhou2025daily} & V+A & \cmark & 30--60 & 42.8 & A\&M & \cmark & \xmark & MC \\
WorldSense~\cite{hong2025worldsense} & V+A & \cmark & 15--656 & 141.1 & M & \cmark & \xmark & MC \\
OmniVideoBench~\cite{li2025omnivideobench} & V+A & \cmark & 4--1955 & 384.2 & M & \cmark & \xmark &  MC \\
JointAVBench~\cite{chao2025jointavbench} & V+A & \cmark & - & 97.2 & A\&M & \cmark & \xmark & MC \\
\midrule
\textbf{LVOmniBench (Ours)} & \textbf{V+A} & \cmark & \textbf{613--5482} & \textbf{2069.7} \textbf{\textcolor{red}{(\smallrocket\ 6$\times$)}} & \textbf{M} & \cmark & \cmark &  \textbf{MC} \\
\bottomrule
\end{tabular}
}
\label{tab:benchmark_compare}
\vspace{-4mm}
\end{table*}

%% file: sec/2_related.tex
\section{Related Work}

\subsection{Omnimodal Large Language Models}

Research on multimodal large language models (MLLMs)~\cite{chen2023vlp,liu2023visual,llava-ov,bai2025qwen2,wang2024qwen2,awadalla2023openflamingo,team2025kimi,ding2025kimi,bai2025qwen3vl,chen2024internvl,team2025gemma,ge2025arc,lin2023video,zhang2025videollama,feng2025efficient} is transitioning from isolated single-modality perception toward omnimodal architectures capable of jointly processing text, images, video, and audio, thereby facilitating practical audio-visual comprehension in real-world scenarios.
This trajectory is exemplified by recent advanced models designed to seamlessly process continuous video and audio streams to generate text and speech outputs~\cite{xu2025qwen2,xu2025qwen3,sun2024video,tang2025video,damonlpsg2024videollama2,fu2025vita,ye2025omnivinci,li2024baichuanomni,chen2025chronusomni,yang2025humanomniv2,tong2025interactiveomni,team2025longcat,ai2025ming,liu2025ola,shu2025audio,sun2025engagement,yao2024minicpm}.
Furthermore, the Gemini series serves as a strong baseline, distinguished by robust omnimodal understanding capabilities~\cite{team2023gemini,team2024gemini,comanici2025gemini}.
Despite the proliferation of SoTA models, the evaluation of audio-visual comprehension predominantly focuses on short video clips or static images.
Our experiments reveal that current models continue to struggle with tasks requiring long-range temporal reasoning, highlighting an inadequacy in processing extended audio-visual inputs.
Consequently, there is an urgent need to develop benchmarks specifically tailored for the comprehension of long audio-visual content.

\subsection{Multimodal Benchmarks}

The rapid advancement of MLLMs has been significantly propelled by evaluation benchmarks.
At present, evaluation benchmarks targeting unimodal comprehension, encompassing isolated image~\cite{li2023seed,hu2025video,antol2015vqa,hudson2019gqa,liu2024mmbench,he2020pathvqa,marino2019ok,lu2023mathvista,fu2023mme,yue2024mmmu,feng2025can,feng2025rewardmap}, video~\cite{fu2025video,wang2025lvbench,li2024mvbench,mangalam2023egoschema,xiao2021next,yu2019activitynet,wu2024longvideobench,zhou2025mlvu,maaz2023video,yuan2025videodeepresearch,song2024moviechat,liu2024tempcompass}, and audio understanding~\cite{panayotov2015librispeech,chen2020vggsound}, are relatively mature.
However, following the emergence of OmniLLMs, evaluating joint audio-visual reasoning poses a significant challenge.
Most existing benchmarks are constrained to domain-specific evaluations or the processing of static images~\cite{yang2022avqa,li2024omnibench,yuan2025videodeepresearch,li2022learning,geng2025longvale}.
Although recent advanced benchmarks, including WorldSense~\cite{hong2025worldsense} and Daily-Omni~\cite{zhou2025daily}, have been proposed, they predominantly focus on short video clips~\cite{zhou2025daily,chao2025jointavbench,nguyen2025see,yang2025audio,sung2024avhbench}.
While OmniVideoBench~\cite{li2025omnivideobench} includes a subset of videos lasting 10 to 30 minutes, the vast majority of the dataset consists of videos lasting only a few minutes.
This brief duration is misaligned with the lengths of videos typically encountered in real-world scenarios.
Consequently, we introduce a new benchmark dedicated to the omnimodal understanding of long-form audio-visual inputs.
The average duration of videos in LVOmniBench exceeds thirty minutes, which is six to twenty times longer than the durations found in previous benchmarks.
Furthermore, the automated generation of questions using LLMs faces challenges in capturing complex, real-world reasoning requirements and remains prone to hallucination~\cite{zhou2025daily,chao2025jointavbench,cao2025xgc}.
To ensure the highest-quality evaluation, all videos and questions in LVOmniBench were manually selected and annotated by human experts.

%% file: sec/3_benchmark.tex
\section{LVOmniBench}
\label{sec:benchmark}

\begin{figure}[t]
    \centering
    \includegraphics[width=1\textwidth]{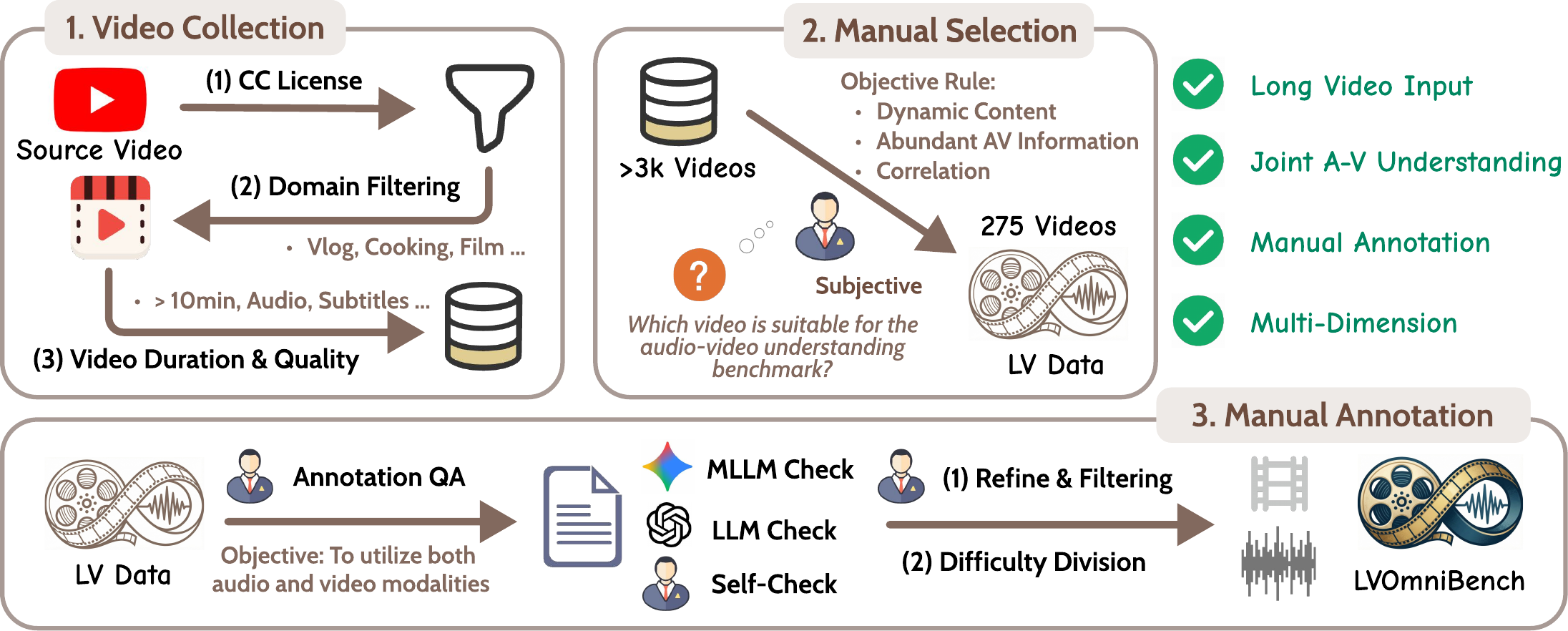}
    \vspace{-5mm}
    \caption{The construction of LVOmniBench follows a rigorous pipeline encompassing video collection, filtering, and question annotation. To guarantee both high-fidelity data quality and sufficient challenge for OmniLLMs, every component, from raw videos to the final questions, underwent meticulous manual selection and annotation.}
    \label{fig:pipline}
    \vspace{-6mm}
\end{figure}

\subsection{Video Collection}
\label{sec:video_c}

\begin{figure}[t]
    \centering
    \includegraphics[width=1\textwidth]{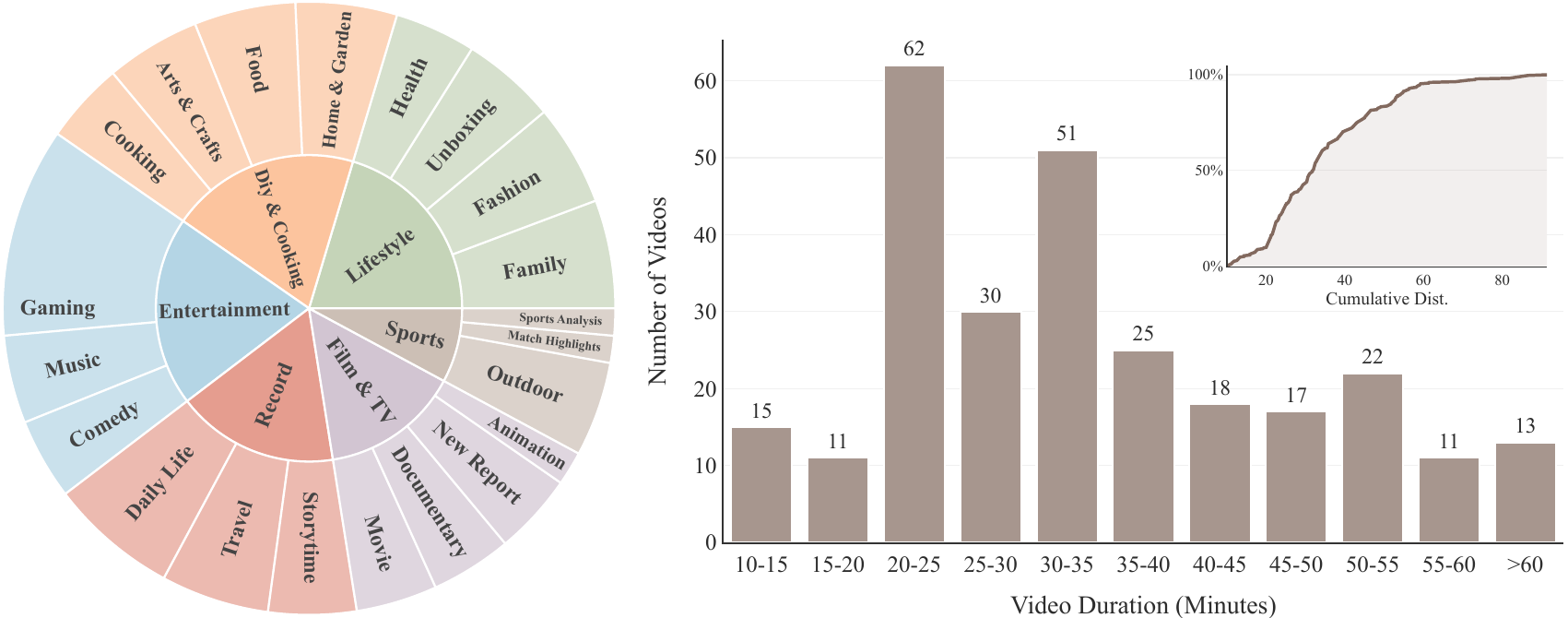}
    \vspace{-6mm}
    \caption{Distribution of videos.
    \textbf{Left:} Videos in LVOmniBench span five primary categories encompassing 21 fine-grained subcategories.
    Each video is selected to ensure sufficient audio-visual information and dynamic variations.
    \textbf{Right:} The durations of the collected videos range from 10 to 90 minutes, with most between 20 and 50 minutes.
    }
    \label{fig:benchmark_video_bar}
    \vspace{-5mm}
\end{figure}

To ensure dynamic audio-visual content and broad thematic coverage, we sourced videos from YouTube to establish a diverse corpus, as shown in \cref{fig:pipline}.
To ensure long-term accessibility and strictly comply with copyright regulations, all videos in LVOmniBench adhere to \emph{Creative Commons} licenses.
This guarantees that the benchmark remains open-source for the research community.
To verify that the videos are well-suited for complex audio-visual reasoning tasks, we systematically categorized them into five broad domains: \emph{entertainment, lifestyle, DIY \& cooking, record, film \& TV.}
This process involved rigorous keyword-based screening and collection across 21 fine-grained subcategories (see \cref{fig:benchmark_video_bar}).

Subsequently, we applied strict length and quality controls, ultimately amassing an initial pool of more than 3,000 raw videos.
Each video ranges from 10 to 90 minutes in duration and features a synchronized audio track.
Unlike video-only benchmarks, not every video satisfies the stringent prerequisites for cross-modal audio-visual reasoning.
Consequently, we meticulously filtered the initial pool to identify dynamic and informative content, curating a final set of 275 high-quality, long videos suitable for question annotation.
As shown in \cref{fig:benchmark_video_bar}, most video durations fall between 20 and 45 minutes, aligning with the typical length distribution of videos in real-world scenarios.

\subsection{Question Answer Annotation}
\label{sec:q_an}
To systematically evaluate long audio-visual comprehension, we first establish a comprehensive taxonomy of question types tailored to the capabilities of OmniLLMs.
These categories aim to assess the proficiency of models in temporal feature alignment, fine-grained understanding, and complementary reasoning across multimodal inputs, including complex scenarios requiring the simultaneous application of multiple cognitive skills.

\begin{figure}[t]
    \centering
    \includegraphics[width=1\textwidth]{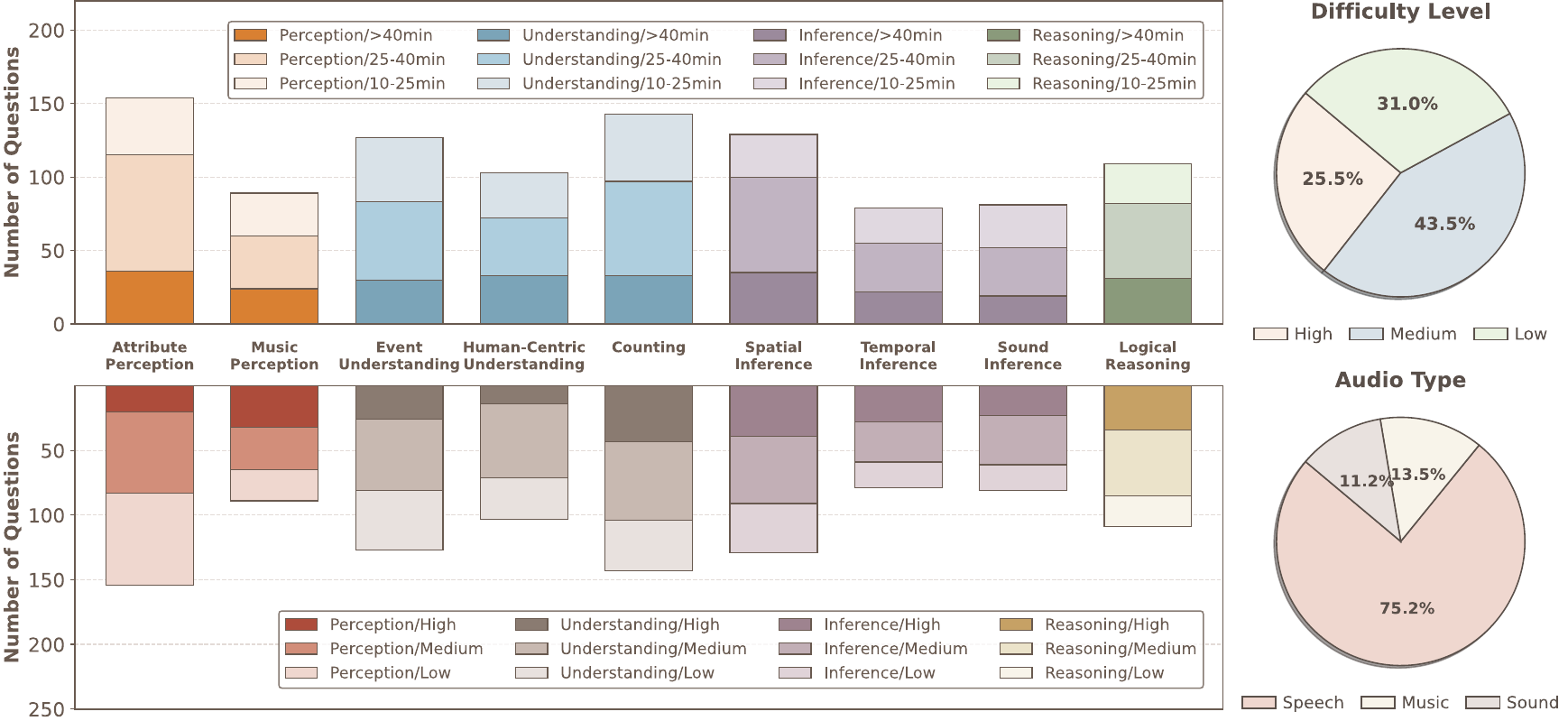}
    \vspace{-5mm}
    \caption{Distribution of questions. 
    \textbf{Left:} LVOmniBench comprises nine question subcategories, with each demonstrating a balanced distribution across difficulty levels and video durations. 
    \textbf{Right:} This panel illustrates the distribution of question difficulty and the corresponding audio types required to answer the questions.}
    \label{fig:benchmark_question_bar}
    \vspace{-6mm}
\end{figure}

\vspace{1mm}
\noindent \textbf{Perception.}
This dimension focuses on extracting multimodal information.
It evaluates the capacity to perceive fundamental acoustic and visual features, including object attributes (e.g., color, texture, shape), quantities, and musical elements.
This layer is crucial for validating the ability of the model to extract fine-grained details from long-context inputs.
This dimension comprises the following subcategories: \emph{Counting, Attribute Perception, and Music Perception.}

\vspace{1mm}
\noindent \textbf{Understanding.}
This category aims to evaluate the proficiency in recognizing entities, actions, and their contextual roles within complex scenes.
Tasks encompass \emph{Human-Centric Understanding} (e.g., identity tracking, emotion recognition, and behavioral modeling) and fine-grained \emph{Event Understanding}.
These evaluations rigorously test the ability of the model to synthesize complementary audio-visual cues across long-term contexts.

\vspace{1mm}
\noindent \textbf{Inference.}
This dimension evaluates the ability to comprehend complex spatiotemporal dynamics and reason about sound events within long-form audio-visual inputs.
Such inference requires rigorous cross-modal alignment across both temporal and spatial dimensions.
Specifically, this category encompasses three distinct subtasks: \emph{Sound Inference}, \emph{Spatial Inference}, and \emph{Temporal Inference}.

\vspace{1mm}
\noindent \textbf{Logical.}
This evaluative dimension necessitates multi-step reasoning, causal tracking, and complex inference grounded in complementary cross-modal information.
Crucially, these tasks cannot be resolved through superficial feature matching or basic audio-visual alignment.
Instead, they require the model to comprehend extended contextual dependencies and construct robust logical reasoning chains across modalities.

After selecting the videos and question types, we annotated the selected high-quality videos using a multiple-choice format.
Specifically, each annotator generated between 1 and 20 questions per video; the exact number scaled according to the duration of the video and the density of relevant audio-visual events.
Each question was formulated with four candidate options.
To ensure the rigor of the benchmark, we enforced strict annotation guidelines:
\textbf{(1)} Questions must necessitate joint audio-visual reasoning to prevent unimodal bias, and the correct option must be unambiguous.
\textbf{(2)} Questions cannot be answered by relying solely on prior commonsense knowledge.
Furthermore, the length of the four options was required to be uniform, and the distractors had to be directly derived from the video or audio.
\textbf{(3)} Annotators were instructed to minimize the use of explicit timestamps in the prompts, ensuring that any temporal references provided did not offer trivial shortcuts to the solution.
Consequently, this first round of annotation yielded more than 1,500 candidate QA pairs.

\subsection{Question Refine and Filtering}
Following the initial annotation phase, we implemented a rigorous evaluation and filtering pipeline.
First, we leveraged the Gemini model to conduct inference testing across three unimodal baselines: video-only, audio-only, and text-only.
Based on the corresponding outputs, we required annotators to refine or delete QA pairs that could be answered \emph{effectively} using a single modality, as this indicates a failure to properly integrate audio-visual cues during annotation.
Furthermore, we utilized the reasoning summaries of the model to systematically filter out questions relying on common sense and flawed designs.
Additionally, we observed that certain annotators depended excessively on timestamps and explicit descriptions, inadvertently introducing unimodal bias and reducing the difficulty of temporal grounding and modality perception.
Consequently, we rigorously refined or discarded such questions.
Finally, after this comprehensive quality screening, we obtained 1,014 QA pairs as the final benchmark dataset.

\input{tab/benchmark_inform}
\vspace{1mm}
\noindent \textbf{Difficulty Level Annotation.}
To provide a more meaningful hierarchical evaluation for the community, we recognize that superficial metrics, such as video duration, question type, or audio modality, are insufficient to accurately reflect the difficulty of a given task or gauge model performance. 
Therefore, we evaluate each QA pair across multiple dimensions, including perceptual difficulty, informational granularity, temporal span, and inference complexity, and stratify the overall difficulty into three tiers: Low, Medium, and High.

\subsection{Dataset Statistics and Comparison}
As detailed in \cref{tab:benchmark_compare,tab:benmark_infor}, the proposed LVOmniBench contains 275 videos spanning five categories and 21 subclasses, with an average duration of 34 minutes and 29 seconds, which is 6-20 times longer than all previous benchmarks~\cite{li2024omnibench,zhou2025daily,hong2025worldsense,li2025omnivideobench,sung2024avhbench,chao2025jointavbench,cao2025xgc,wang2025lvbench}.
The dataset comprises 1,014 multiple-choice questions across nine categories, with an average question length of 16.4 words.
As illustrated in \cref{fig:benchmark_question_bar}, the distribution of requisite audio types for answering the questions, specifically speech, music, and sound, is 763:137:114, respectively, which reflects the prevalence of speech-driven interactions in real-world scenarios.
Regarding the difficulty gradient, the distribution of low, medium, and high questions is 315:441:259, with all three tiers represented across all question categories.
Overall, to the best of our knowledge, LVOmniBench is the first benchmark to comprehensively evaluate the comprehension of OmniLLMs in long-form audio-visual scenarios.
We aim to catalyze future advancements in processing extended context lengths and joint audio-visual inputs.
To this end, our benchmark establishes a robust foundation for the comprehensive assessment and in-depth analysis of OmniLLM capabilities on prolonged multimodal sequences.

%% file: tab/benchmark_inform.tex
\begin{table}[t]
\centering
\setlength{\tabcolsep}{10pt}
\caption{Key statistics of our LVOmniBench benchmark.}
\vspace{-2mm}
\label{tab:benchmark_stats_folded}
\resizebox{0.85\linewidth}{!}{%
\begin{tabular}{l c @{\hspace{8mm}} l c}
\toprule
\multicolumn{2}{l}{Video Information} &  \multicolumn{2}{l}{Question Information} \\
\midrule
Total Video & 275 & Total Questions & 1014  \\
Categories & 21 & Question Type & 9 \\
Audio Types & Sp+So+Mu & Difficulties (Low:Med:High) & 314:441:259 \\
Duration & 613–5482 & Avg Question Length & 16.4 \\
Avg. Duration  & 34:29 &  Multiple-Choice & 4 (A/B/C/D)\\
\bottomrule
\end{tabular}%
}
\label{tab:benmark_infor}
\vspace{-4mm}
\end{table}

%% file: sec/4_exp.tex
\section{Experimental Results}
\label{sec:exp}
\input{tab/main}
\subsection{Experimental Settings}

\noindent \textbf{Models.} To provide a comprehensive evaluation of OmniLLM performance on long-form audio-visual inputs, we benchmark several leading open-source models: Ming-Flash-Omni-2.0-100B~\cite{ai2025ming}, MiniCPM-o 4.5~\cite{yao2024minicpm}, Qwen3-Omni-30B~\cite{xu2025qwen3}, video-SALMONN 2+ 7B~\cite{tang2025video}, Qwen2.5-Omni-7B~\cite{xu2025qwen2}, and VideoLLaMA2-7B~\cite{damonlpsg2024videollama2}. 
Furthermore, we evaluate the performance of the Video LLMs Qwen3-VL-8B and Qwen3-VL-30B~\cite{bai2025qwen3vl}, alongside the Audio LLMs Qwen2-Audio~\cite{chu2024qwen2}. 
Finally, we incorporate the Gemini~\cite{team2023gemini,team2024gemini,comanici2025gemini} as a robust, proprietary baseline, leveraging its SoTA omnimodal comprehension capabilities.

\vspace{1mm}
\noindent \textbf{Implementation Details.}
We use the official configurations for each model and try to evaluate using the maximum permissible number of frames. 
For Qwen2.5-Omni, Qwen3-Omni, Qwen3-VL, and video-SALMONN 2+, we set the number of input frames to 768, ensuring maximal utilization of the model's context length without exceeding architectural limits. 
Conversely, due to stricter context length limitations, we restricted the number of input frames for MiniCPM-o 4.5 and VideoLLaMA2-7B to 64 and 16, respectively.
All local experiments were conducted using NVIDIA H100 (80GB) and L40S (48GB) GPUs.
For the Gemini 2.0 and Gemini 3.0 series models, we set the input frame rate to 1 frame per second (FPS), with the Gemini series specifically configured to utilize its deep thinking mode.

\begin{figure*}[t]
    \centering
    \begin{subfigure}[t]{0.48\linewidth}
        \centering
        \captionsetup{font={footnotesize}, skip=4pt}
        \includegraphics[width=1\linewidth]{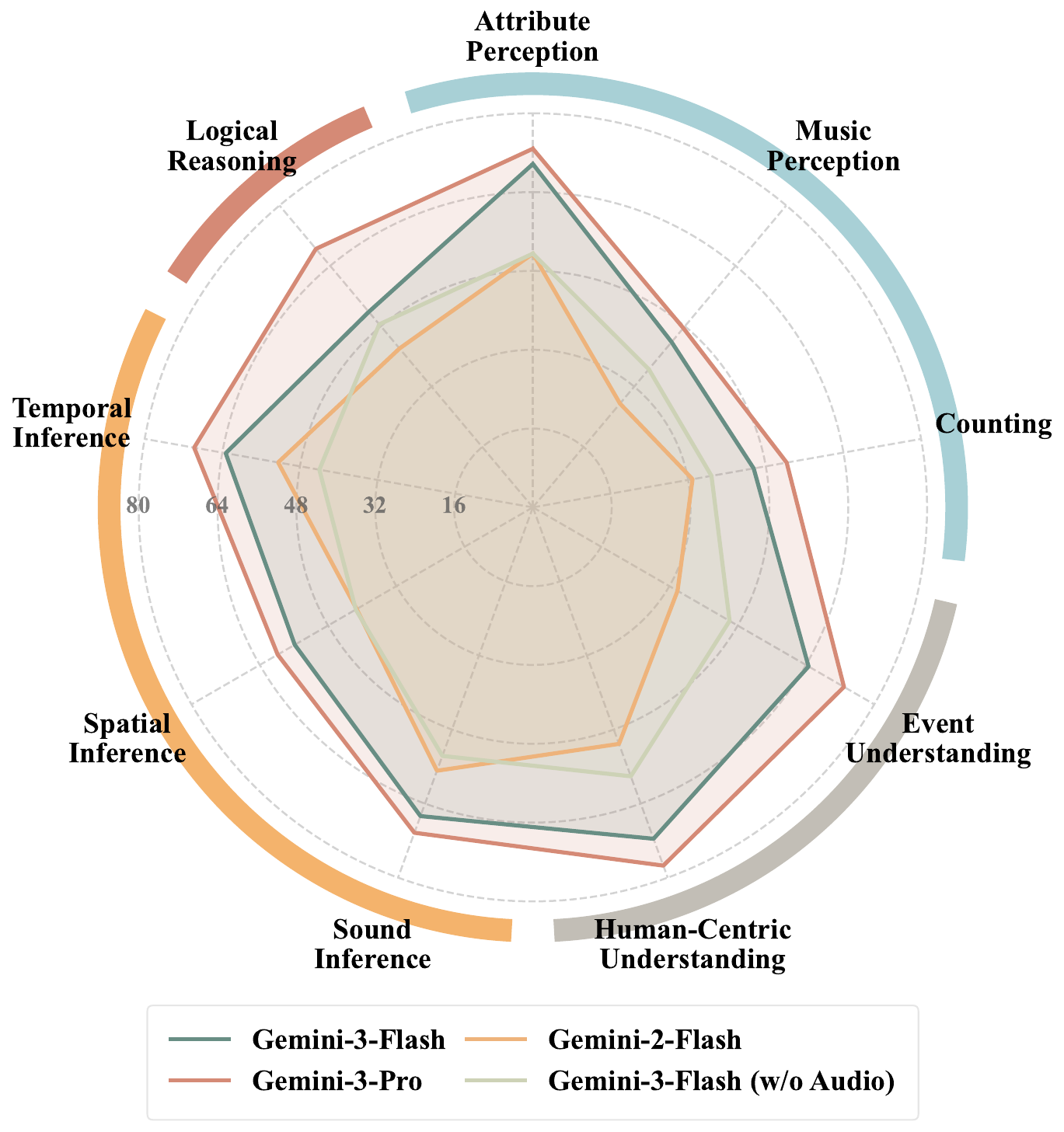}
    \end{subfigure}%
    \hspace{3mm}
    \begin{subfigure}[t]{0.48\linewidth}
        \centering
        \captionsetup{font={footnotesize}, skip=4pt}
        \includegraphics[width=1\linewidth]{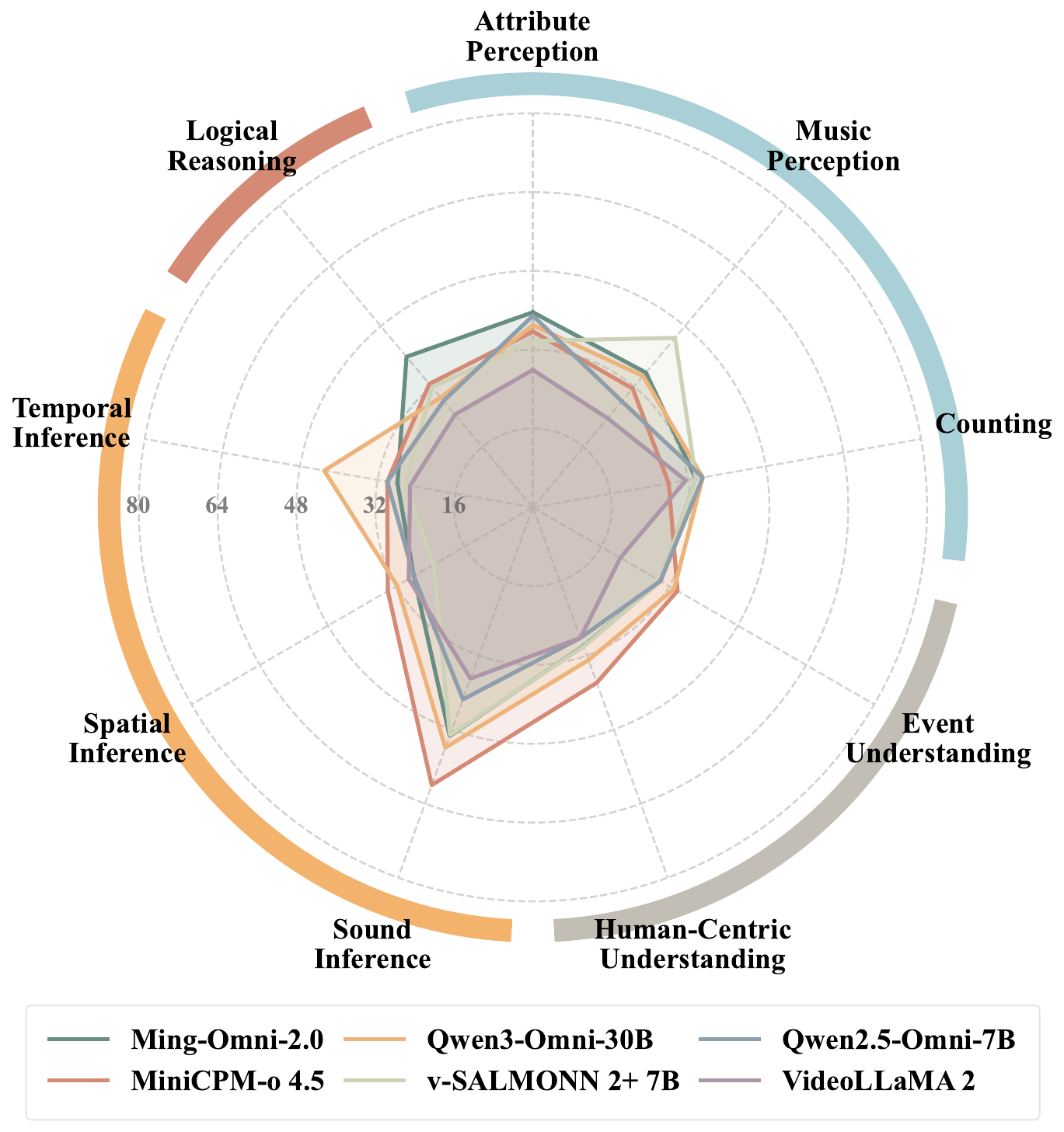}
    \end{subfigure}%
    \caption{Comparison between proprietary models and open-source models on LVOmniBench across different tasks. 
    The blue, red, orange, and gray in the outer
circle stands for perception, logical, inference, and understanding tasks. As can be observed, proprietary models demonstrate a substantial performance advantage over their open-source counterparts. Furthermore, across all models, significant vulnerabilities remain in specific sub-tasks, notably Counting and Music Perception.}
    \label{fig:radar}
\vspace{-2mm}
\end{figure*}

\subsection{Quantitative Results}
\label{sec:main_results}
\begin{figure}[t]
    \centering
    \includegraphics[width=1\textwidth]{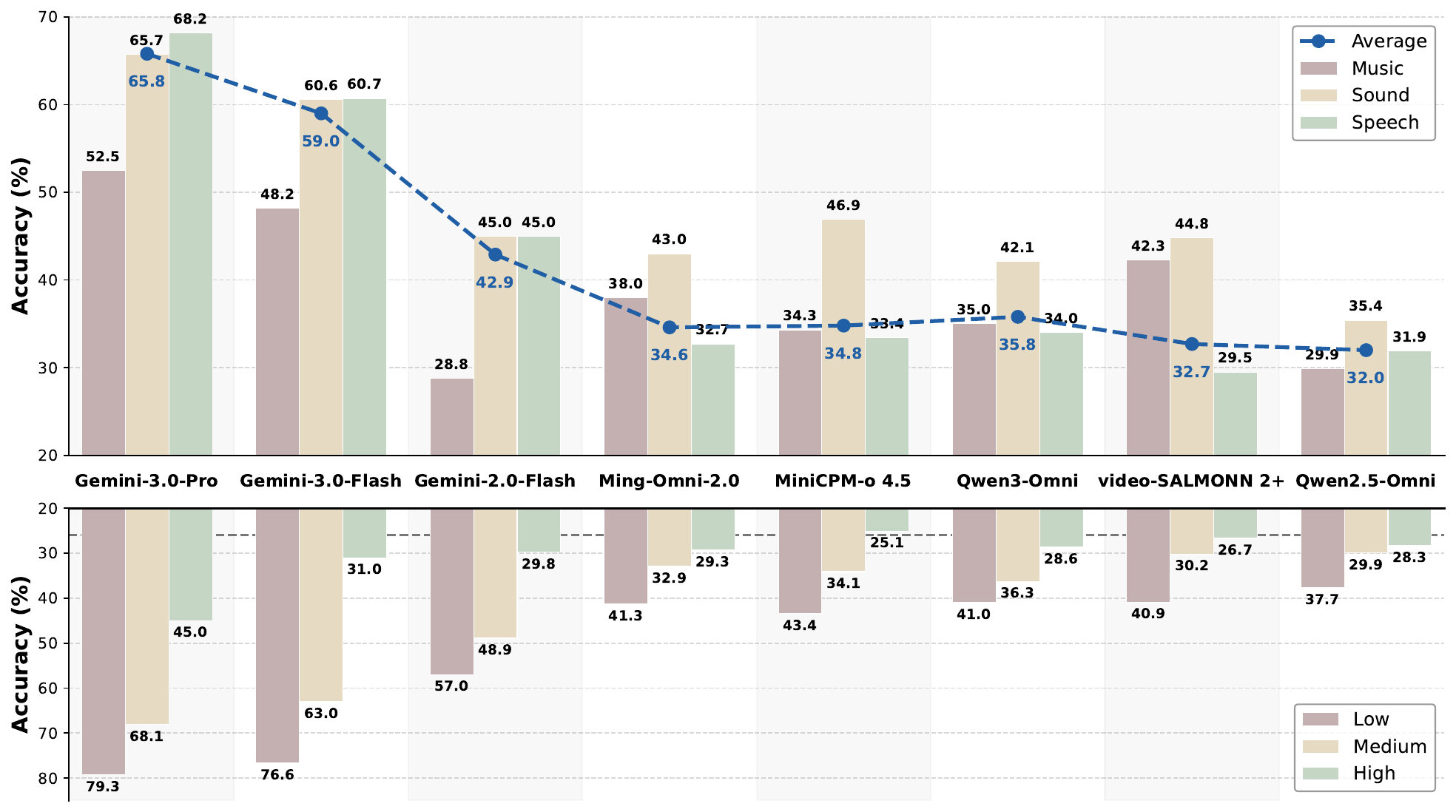}
    \vspace{-6mm}
    \caption{Results across various audio categories and difficulty levels.
    The findings demonstrate that proprietary models consistently outperform open-source models across all evaluated dimensions.
    Furthermore, OmniLLMs generally face greater challenges in perceiving musical audio; specifically, at high difficulty levels, the performance of open-source models declines to a level approaching random chance.}
    \label{fig:results_audio_type}
    \vspace{-7mm}
\end{figure}

\vspace{1mm}
\noindent \textbf{Performance of Proprietary Models.}
As shown in \cref{tab:main}, the Gemini 3.0 series, currently the leading proprietary architecture in the field of omnimodal comprehension, achieves the highest overall accuracy on LVOmniBench.
Specifically, the Flash and Pro variants achieve accuracies of 59.0\% and 65.8\%, respectively, representing a 1.5-fold improvement over the Gemini 2.0 Flash.
This superior performance is attributable to the capacity of these models to process ultra-long video contexts, alongside robust audio comprehension and precise temporal alignment.
Analyzing performance across difficulty tiers, the overall distribution of accuracy closely aligns with our annotated difficulty gradients; notably, Gemini 3.0 Pro maintains an accuracy of 45\% even on high-difficulty questions.
This consistency validates the effectiveness of our benchmark for rigorous hierarchical evaluation.
Furthermore, as shown in \cref{fig:radar}, a granular analysis categorized by question type reveals that tasks requiring counting and music perception remain exceptionally challenging.
Finally, as depicted in \cref{fig:results_audio_type}, the capacity of models to perceive non-verbal and abstract sounds within the music category is significantly lower than the performance of these models on the other two audio types.
These findings highlight that robust cross-modal alignment for non-linguistic audio classes remains an urgent challenge.

\vspace{1mm}
\noindent \textbf{Performance of Open-Source Models.}
Qwen3-Omni achieves the highest accuracy of 35.8\%, while all other open-source models fall below the 35\% threshold.
As detailed in \cref{tab:main}, this starkly illustrates that current open-source architectures struggle to process and comprehend long-form audio-visual inputs effectively.
On high-difficulty tasks, the performance of these models degrades to near-random chance.
Furthermore, \cref{fig:radar} highlights significant performance variance across distinct task categories.
For instance, Qwen3-Omni excels in temporal inference, Ming-Flash-Omni-2.0 demonstrates superior logical reasoning capabilities, and MiniCPM-o 4.5 exhibits relatively better audio perception.
Additionally, we observe that accuracy scores in the range of 30\% to 35\% largely reflect models relying on superficial visual priors rather than genuinely achieving cross-modal comprehension.
The severe computational overhead introduced by the processing of extended temporal contexts remains a primary bottleneck limiting the capabilities of open-source models.
Consequently, OmniLLMs still require substantial optimization to bridge this performance gap.

\vspace{1mm}
\noindent \textbf{Effectiveness of Omnimodal Understanding.}
The performance of the unimodal baselines, specifically Qwen3-Omni (audio-only) and Qwen2-Audio, underscores the necessity of joint visual and audio comprehension within LVOmniBench, as illustrated in \cref{tab:main} and \cref{tab:audio_video_use}(a).
In addition, the video-only performance of models on LVOmniBench was evaluated; for instance, Qwen3-VL, despite its specialized optimization for long-video understanding~\cite{bai2025qwen3vl,fu2025video,wang2025lvbench,wu2024longvideobench}, achieves only 36\% accuracy.
This performance represents a substantial decline relative to its results on prior benchmarks that focused exclusively on long-video understanding.
Furthermore, as detailed in \cref{tab:main,tab:audio_video_use}, restricting Gemini 3 Flash to visual-only reasoning results in a significant accuracy reduction of up to 13\%; this ablation study clearly validates the essential role of audio cues in the proposed benchmark.

\subsection{Key Takeaways}
\input{tab/results_with_audio}

\noindent \textbf{Decisive Lead of Proprietary Models.}
As indicated in \cref{tab:main} and \cref{fig:radar,fig:results_audio_type}, proprietary models consistently outperform open-source baselines across all facets of long-form audio-visual comprehension, spanning both individual tasks and various audio categories.
This divergence highlights a critical distinction from video-only comprehension.
While advanced methodologies and frameworks have rendered open-source models highly competitive in purely long-video tasks, the inherent \emph{complexity of joint audio-visual processing} and the rigorous demands of \emph{fine-grained cross-modal alignment} reveal a substantial performance gap.
Consequently, it is expected that LVOmniBench will catalyze future research dedicated to addressing these cross-modal challenges.

\vspace{1mm}
\noindent \textbf{Enhancing Cross-Modal Capabilities of OmniLLMs.}
As demonstrated in \cref{sec:main_results}, open-source OmniLLMs exhibit performance levels near random chance on LVOmniBench.
\cref{tab:audio_video_use} provides a detailed performance breakdown across a suite of OmniLLMs.
It is observed that most models have challenges in effectively utilizing or comprehending the audio modality, raising a critical question:

\textbf{\emph{Can the open-source models genuinely use the audio in the video?}}

\noindent The analysis suggests \textbf{\underline{no}}.
To rigorously investigate this hypothesis, the audio tracks were transcribed using Gemini 2.5 to obtain automatic speech recognition (ASR) transcripts.
By providing this ASR text to the model as an explicit supplementary cue alongside the visual stream, significant improvements were observed for Qwen3-Omni and Ming-Flash-Omni-2.0.
These results indicate that the inability of open-source models to leverage extended audio information effectively drives the observed decrease in accuracy.
\begin{highlightbox}
Overall, achieving cross-modal alignment between audio and visual within OmniLLMs remains a formidable challenge, representing a critical bottleneck that warrants further exploration.
\end{highlightbox}

\vspace{1mm}
\noindent \textbf{Impact of Audio Input.}
In contrast, Gemini 3.0 Flash demonstrates robust multimodal processing; supplementing the visual stream with ASR text does not yield improvements in accuracy over the original raw audio.
This suggests that raw audio encapsulates critical non-verbal signals, such as emotion, intonation, and musicality, that extend beyond mere textual semantics.
These findings underscore the need to preserve comprehensive acoustic cues for better omnimodal understanding in real-world video scenarios.
This discrepancy not only highlights the superior performance of proprietary models but also validates the \emph{effectiveness} of our benchmark design.

\vspace{1mm}
\noindent \textbf{Long Audio-Video Understanding Capabilities.}
We observe that as the duration of audio and video inputs increases, model performance degrades significantly compared to that on short-video benchmarks~\cite{zhou2025daily,hong2025worldsense}.
In the context of long-form video understanding, numerous methodologies have been proposed to mitigate the information bottleneck posed by processing comprehensive video content with a limited number of input frames.
These approaches encompass training-free token compression strategies~\cite{tao2025dycoke,shao2025holitom,liu2025video,shao2025tokens,shen2025fastvid,huang2024prunevid}, dynamic input resolutions~\cite{bai2025qwen2,bai2025qwen3vl}, streaming understanding~\cite{di2025rekv,xu2025streamingvlm,timechatonline,chen2025streamingtom,wang2025accelerating}, and specialized model architectures~\cite{shu2024video,liu2025video,shen2024longvu} and agents~\cite{zhang2025deep,zhang2024omagent,fan2024videoagent,wang2024videoagent,yang2025vca,wang2025videotree,tian2025ego,ding2025videozoomer} tailored for long videos.
However, within the realm of joint audio-visual comprehension, this bottleneck becomes particularly pronounced~\cite{tao2025omnizip,ding2026omnisift}.
As an extended temporal sequence, audio inherently possesses a high information density and exhibits strict sequential dependencies.
Consequently, we believe that the primary limitation in long-form audio-visual comprehension currently stems from the challenges of modeling long audio sequences.
Therefore, enhancing the capabilities for long-form audio-visual modeling is imperative for the next generation of OmniLLMs to effectively comprehend extended-duration real-world video inputs.

\subsection{Error Analysis}
\begin{wrapfigure}{r}{0.5\textwidth}
\centering
\captionsetup{font={small}, skip=2pt}
\vspace{-10mm}
\includegraphics[width=0.5\textwidth]{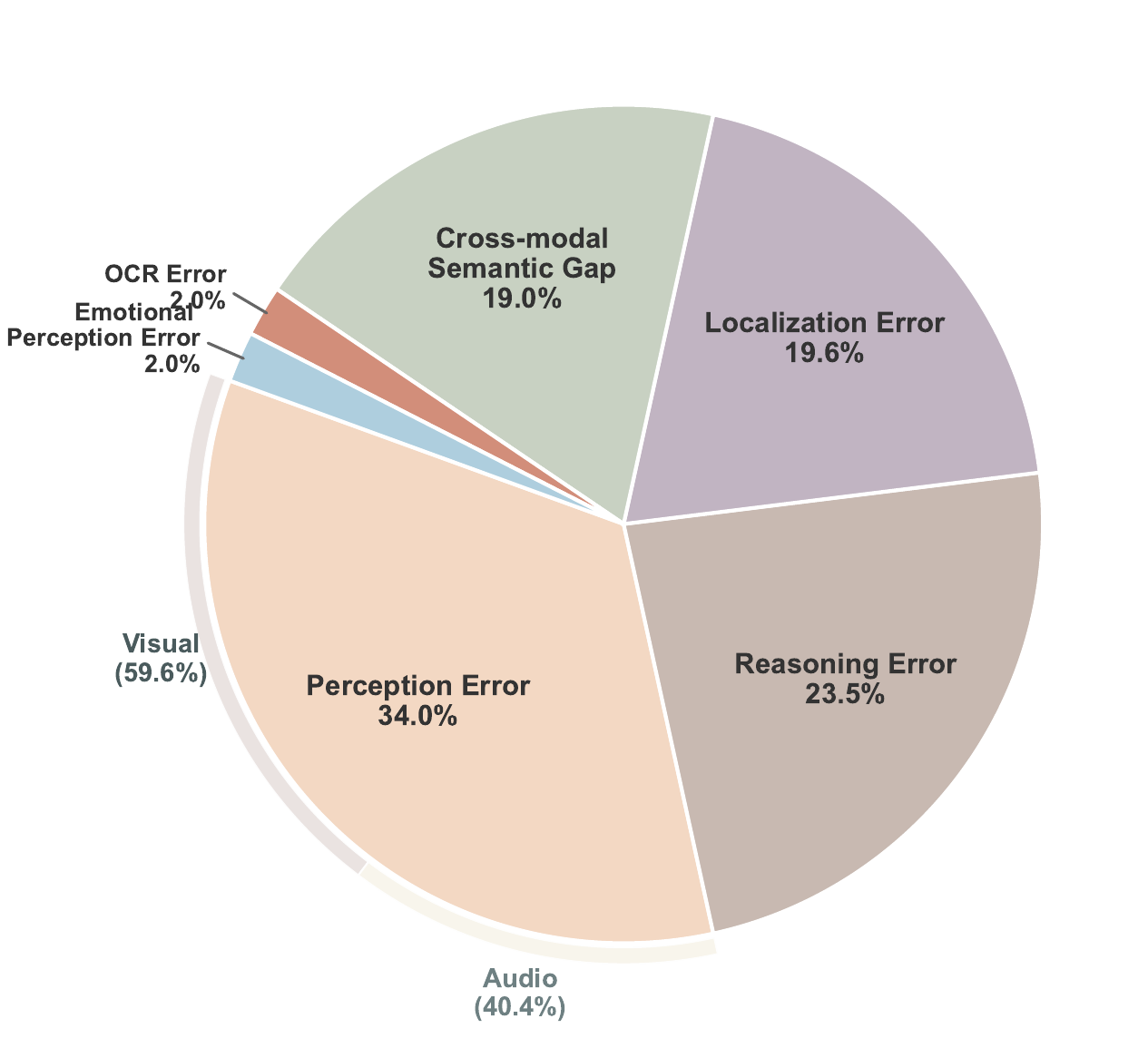}
\vspace{-5mm}
\caption{Error distribution of six types in 153 annotated Gemini 3 Flash errors of LVOmniBench. Perception errors were divided into audio and visual categories, accounting for 60\% and 40\%, respectively.}
\vspace{-6mm}
\label{fig:data_error}
\end{wrapfigure}
In this section, we aim to provide an in-depth analysis of model predictions and failure modes to inform future research directions. 
Generally, OmniLLMs exhibit notable limitations in interpreting abstract audio representations, such as music, and struggle with tasks requiring precise counting or spatial reasoning. 
We manually analyzed 153 incorrect predictions from Gemini 3 Flash, categorizing the root causes to provide a comprehensive statistical breakdown and discussion.

\vspace{1mm}
\noindent \textbf{Perception Errors (34\%).} Perception errors are broadly categorized into audio and visual modalities, as illustrated in \cref{fig:data_error}.
This categorization assesses the ability of models to comprehend fundamental audio and video information.
Concerning audio perception failures, we observe a prevailing modality bias: models disproportionately rely on visual cues while \emph{neglecting corresponding audio signals}, which ultimately results in incorrect predictions.
Furthermore, accurately interpreting fine-grained acoustic properties, such as sound intensity, timbre, and pitch, poses a significant challenge for current models.
Conversely, visual perception errors predominantly stem from a lack of spatial reasoning and detailed comprehension, as models frequently struggle to identify object counts, spatial directions, and visual attributes.

\vspace{1mm}
\noindent \textbf{Localization Errors (19.6\%).} Given that long-form video and audio inputs span tens of minutes, precise temporal grounding is paramount.
While numerous prior studies have focused extensively on video-only temporal grounding~\cite{wang2025time,wu2025survey,zhang2025timelens,zeng2024timesuite,qu2024chatvtg,huang2024vtimellm}, the joint localization of audio and video modalities remains largely underexplored.
By highlighting this critical gap, we aim to inspire future research toward unified omnimodal temporal localization.

\vspace{1mm}
\noindent \textbf{Cross-Modal Semantic Gap (19\%).} Models frequently process the two modalities in isolation, which makes it highly challenging to fuse and align their complementary information in a manner consistent with human cognition.
Although models can often independently comprehend isolated audio and video streams, they struggle with cross-modal correspondence.
Consequently, they exhibit suboptimal performance in tasks that require fine-grained alignment, such as human-centered understanding and attribute perception.

\vspace{1mm}
\noindent \textbf{Reasoning Errors (23.5\%).}
Even when a model successfully perceives both modalities and their cross-modal alignment, it frequently struggles with logical deduction and mathematical reasoning. 
Furthermore, deciphering complex temporal dynamics, such as action sequencing and event causality, alongside spatial relationships extracted from audio-visual cues, remains a significant challenge. 
Consequently, moving beyond fundamental perception and enhancing better reasoning capabilities is critical to addressing these limitations.

\vspace{1mm}
\noindent \textbf{Other Errors and Summary.}
The remaining error categories encompass OCR (2\%) and emotional perception errors (2\%), stemming from the limited capacity of models to extract embedded textual data and to decipher implicit affective cues within audio streams.
Detailed examples are provided in the Appendix.
\begin{highlightbox}
Overall, OmniLLMs encounter significantly greater bottlenecks in long-form audio-visual comprehension compared to those observed in unimodal or short-video benchmarks.
Through this exhaustive analysis, we aim to inspire future architectural and algorithmic enhancements for OmniLLMs.
\end{highlightbox}

%% file: tab/main.tex
\begin{table*}[t]
\centering
\caption{The performance of models on LVOmniBench. 
We evaluate OmniLLMs, utilizing both Video LLMs and Audio LLMs as specialized baselines. 
During inference, each model is evaluated using the maximum permissible number of input frames. 
The results demonstrate that long-form audio-visual comprehension remains a significant challenge for current SoTA models.}
\vspace{-2mm}
\setlength{\tabcolsep}{5pt}
\resizebox{\linewidth}{!}{
\begin{tabular}{lccccccccc}
\toprule
\multicolumn{1}{l|}{\multirow{2}{*}{Model}} & \multicolumn{1}{c|}{\multirow{2}{*}{Modality}} & \multicolumn{3}{c|}{Difficulty} & \multirow{2}{*}{Understanding} & \multirow{2}{*}{Perception} & \multirow{2}{*}{Inference} & \multicolumn{1}{c|}{\multirow{2}{*}{Logical}} & \multirow{2}{*}{Avg.} \\
\multicolumn{1}{l|}{} & \multicolumn{1}{c|}{} & Low & Medium & \multicolumn{1}{c|}{High} & & & & \multicolumn{1}{c|}{} & \\ \midrule
\rowcolor[gray]{0.95}
\multicolumn{10}{c}{\emph{Proprietary Omnimodal LLMs}} \\
\midrule
\multicolumn{1}{l|}{Gemini-3.0-Pro} & \multicolumn{1}{c|}{A + V} & \textbf{79.3} & \textbf{68.1} & \multicolumn{1}{c|}{\textbf{45.0}} & \textbf{73.3} & \textbf{60.1} & \textbf{65.4} & \multicolumn{1}{c|}{\textbf{67.5}} & \textbf{65.8} \\
\multicolumn{1}{l|}{Gemini-3.0-Flash} & \multicolumn{1}{c|}{A + V} & \underline{76.6} & \underline{63.0} & \multicolumn{1}{c|}{\underline{31.0}} & \underline{67.7} & \underline{54.7} & \underline{60.6} & \multicolumn{1}{c|}{\underline{51.8}} & \underline{59.0} \\
\multicolumn{1}{l|}{Gemini-2.0-Flash} & \multicolumn{1}{c|}{ A + V } & 57.0 & 48.9 & \multicolumn{1}{c|}{29.8} & 41.4 & 38.5 & 49.1 & \multicolumn{1}{c|}{42.1} & 42.9 \\
\rowcolor{mycol1}
\multicolumn{1}{l|}{Gemini-3.0-Flash} & \multicolumn{1}{c|}{V} & 55.6 & 49.3 & \multicolumn{1}{c|}{30.6} & 51.4 & 42.9 & 42.3 & \multicolumn{1}{c|}{48.5} & 46.2 \\
\midrule
\rowcolor[gray]{0.95}
\multicolumn{10}{c}{\emph{Open-source Omnimodal LLMs}} \\
\midrule
\multicolumn{1}{l|}{Ming-Omni-2.0-100B} & \multicolumn{1}{c|}{A + V} & 41.3 & 32.9 & \multicolumn{1}{c|}{29.3} & 30.0 & 36.5 & 33.9 & \multicolumn{1}{c|}{39.1} & 34.6\\
\multicolumn{1}{l|}{MiniCPM-o 4.5} & \multicolumn{1}{c|}{A + V} & 43.4 & 34.1 & \multicolumn{1}{c|}{25.1} & 35.7 & 31.9 & 39.1 & \multicolumn{1}{c|}{32.7} & 34.8\\
\multicolumn{1}{l|}{Qwen3-Omni-30B} & \multicolumn{1}{c|}{A + V} & 41.0 & 36.3 & \multicolumn{1}{c|}{28.6} & 33.0 & 35.8 & 40.5 & \multicolumn{1}{c|}{29.1} & 35.8 \\
\multicolumn{1}{l|}{video-SALMONN 2+ 7B} & \multicolumn{1}{c|}{A + V} & 40.9 & 30.2 & \multicolumn{1}{c|}{26.7} & 30.0 & 36.3 & 30.3 & \multicolumn{1}{c|}{31.8} & 32.7 \\
\multicolumn{1}{l|}{Qwen2.5-Omni-7B} & \multicolumn{1}{c|}{A + V} & 37.7 & 29.9 & \multicolumn{1}{c|}{28.3} & 29.1 & 34.9 & 31.8 & \multicolumn{1}{c|}{28.2} & 32.0 \\
\multicolumn{1}{l|}{VideoLLaMA2-7B} & \multicolumn{1}{c|}{A + V} & 27.0 & 26.8 & \multicolumn{1}{c|}{28.2} & 23.9 & 28.2 & 29.4 & \multicolumn{1}{c|}{24.6} & 27.2\\
\midrule
\rowcolor[gray]{0.95}
\multicolumn{10}{c}{\emph{Video LLMs (Visual)}} \\ 
\midrule
\rowcolor{mycol1}
\multicolumn{1}{l|}{Qwen3-VL-30B} & \multicolumn{1}{c|}{V} & 42.9 & 35.2 & \multicolumn{1}{c|}{30.1} &37.4 & 39.9& 32.5& \multicolumn{1}{c|}{30.9} & 36.3\\
\rowcolor{mycol1}
\multicolumn{1}{l|}{Qwen3-VL-8B} & \multicolumn{1}{c|}{V} & 37.1 & 36.5 & \multicolumn{1}{c|}{32.1} & 32.2 & 37.1 & 36.7 & \multicolumn{1}{c|}{34.6} & 35.6\\
\midrule
\rowcolor[gray]{0.95}
\multicolumn{10}{c}{\emph{Audio LLMs (Audio)}} \\ 
\midrule
\rowcolor{mycol1}
\multicolumn{1}{l|}{Qwen2-Audio} & \multicolumn{1}{c|}{A} & 27.0 & 25.2 & \multicolumn{1}{c|}{21.2} & 25.2& 22.0 & 26.3 & \multicolumn{1}{c|}{29.1} & 24.7\\
\bottomrule
\end{tabular}
}
\label{tab:main}
\vspace{-5mm}
\end{table*}

%% file: tab/results_with_audio.tex
\begin{table*}[t]
\centering
\caption{The impact of audio and video information. 
\textbf{(a)} The evaluation of performance across various capabilities. 
The abbreviations are defined as follows: \textbf{AP}: Attribute Perception, \textbf{EU}: Event Understanding, \textbf{HU}: Human-Centric Understanding, \textbf{LR}: Logical Reasoning, \textbf{MP}: Music Perception, \textbf{SI}: Sound Inference, \textbf{SpI}: Spatial Inference, and \textbf{TI}: Temporal Inference. 
\textbf{(b)} Experiments are conducted across three input configurations: video-only, video with ASR, and video with original audio.}
{
\setlength{\tabcolsep}{7pt}
\resizebox{\linewidth}{!}{
\begin{tabular}{l c ccccccccc c}
    \toprule
    Model & Modality & \makecell{AP} & Counting & \makecell{EU} & \makecell{HU} & \makecell{LR} & \makecell{MP} & \makecell{SI} & \makecell{SpI} & \makecell{TI} & Avg. \\
    \midrule
    Qwen3-Omni-30B   & \makecell{Audio \\ +Video} & \makecell{27.9 \\ 37.0} & \makecell{23.1 \\ 35.0} & \makecell{24.4 \\ 33.1} & \makecell{26.2 \\ 33.0} & \makecell{28.2 \\ 39.1} & \makecell{36.1 \\ 34.8} & \makecell{35.8 \\ 51.9} & \makecell{26.4 \\ 31.8} & \makecell{31.7 \\ 43.1} & \makecell{28.1 \\ $35.8_{+7.7}$} \\
    \midrule
    \rowcolor{mycol2!40!white}
    Gemini-3.0-Flash & A + V & 69.7 & 45.5 & 64.6 & 71.6 & 51.8 & 43.8 & 66.7 & 55.8 & 63.8 & 59.0 \\
    \bottomrule
\end{tabular}
}
}
\\[1mm]
{\small \textbf{(a)} Impact of Video Information}
\\[2mm]
\resizebox{\linewidth}{!}{
\begin{tabular}{lcccccccccccc}
\toprule
\multirow{2}{*}{Model} & \multicolumn{3}{c}{\textbf{Speech}} & \multicolumn{3}{c}{\textbf{Sound}} & \multicolumn{3}{c}{\textbf{Music}} & \multicolumn{3}{c}{\textbf{Overall}} \\
\cmidrule(lr){2-4} \cmidrule(lr){5-7} \cmidrule(lr){8-10} \cmidrule(lr){11-13}
&\multicolumn{1}{c|}{Video} & \multicolumn{1}{c|}{+ASR} & +Audio &\multicolumn{1}{c|}{Video} & \multicolumn{1}{c|}{+ASR} & +Audio & \multicolumn{1}{c|}{Video} & \multicolumn{1}{c|}{+ASR} & +Audio & \multicolumn{1}{c|}{Video} & \multicolumn{1}{c|}{+ASR} & +Audio \\
\midrule
Gemini-3-Flash & 46.5 & $54.2_{+7.7}$ & $60.7_{+14.2}$ & 53.5 & $48.8_{-4.7}$ & $60.6_{+7.1}$ & 38.8 & $44.0_{+5.2}$ & $48.2_{+9.4}$ & 46.2 & $52.0_{+5.8}$ & $59.0_{+12.8}$ \\
Ming-Omni-2.0-100B & 32.6 & $40.8_{+8.2}$ & $32.7_{+0.1}$ & 35.1 & $44.8_{+9.7}$ & $43.0_{+7.9}$ & 37.9 & $36.0_{-1.9}$ & $38.0_{+0.1}$ & 33.6 & $40.5_{+6.9}$ & $34.6_{+1.0}$\\
MiniCPM-o 4.5 & 32.8 & $33.7_{+0.9}$ & $33.4_{+0.6}$ & 35.4 & $28.6_{-6.8}$ & $46.9_{+11.5}$ & 32.8 & $27.1_{-5.7}$ & $34.3_{+1.5}$ & 33.1 & $32.3_{-0.8}$ & $34.8_{+1.7}$ \\
Qwen3-Omni-30B & 34.4 & $42.3_{+7.9}$ & $35.0_{+0.6}$ & 34.2 & $50.1_{+15.9}$ & $42.1_{+7.9}$ & 40.1 & $36.0_{-4.1}$ & $35.0_{-5.1}$ & 35.1 & $42.2_{+7.1}$ & $35.8_{+0.7}$ \\
VideoLLaMA2-7B & 27.7 & -- & $27.1_{-0.6}$ & 37.7 & -- & $31.6_{-6.1}$ & 24.1 & -- & $24.1_{+0.0}$ & 28.4 & -- & $27.2_{-1.2}$ \\
\bottomrule
\end{tabular}
}
\\[1mm]
{\small \textbf{(b)} Impact of Audio Information}

\label{tab:audio_video_use}
\vspace{-5mm}
\end{table*}

%% file: sec/5_conclusion.tex
\section{Conclusion}

In this paper, we introduce LVOmniBench, a pioneering benchmark designed for long audio-video understanding. 
The benchmark features diverse long-form videos with dynamic multimodal content and manually annotated QA pairs, aiming to comprehensively evaluate the capabilities of OmniLLMs in complex joint audio-visual understanding. 
Extensive evaluations reveal a substantial performance disparity between proprietary and open-source models, underscoring the need for advanced architectures capable of jointly processing extended audio and video sequences. 
We hope that LVOmniBench will catalyze future research and drive the development of more robust and comprehensive OmniLLMs.

%% file: sec/supp.tex
\clearpage

\appendix
\section*{Supplementary Material}

\section{Error Analysis Cases}
\begin{figure}[h]
    \centering
    \vspace{-2mm}
    \includegraphics[width=\linewidth]{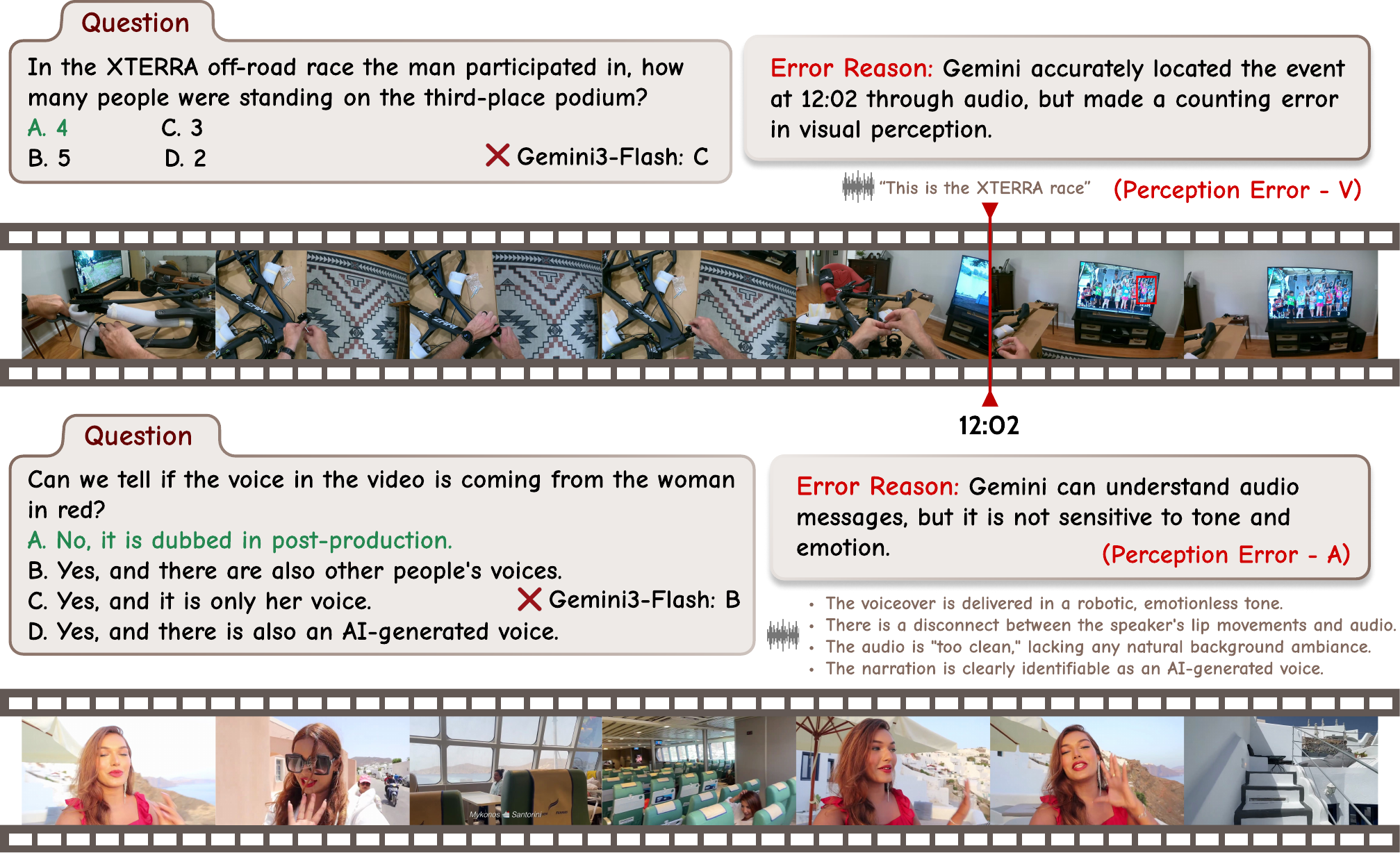}
    \caption{Cases of perception errors. 
    We present examples of failures that arise from the misperception of underlying visual or audio information by models.}
\label{fig:error_per}
\vspace{-6mm}
\end{figure}

\begin{figure}[t]
    \centering
    \includegraphics[width=\linewidth]{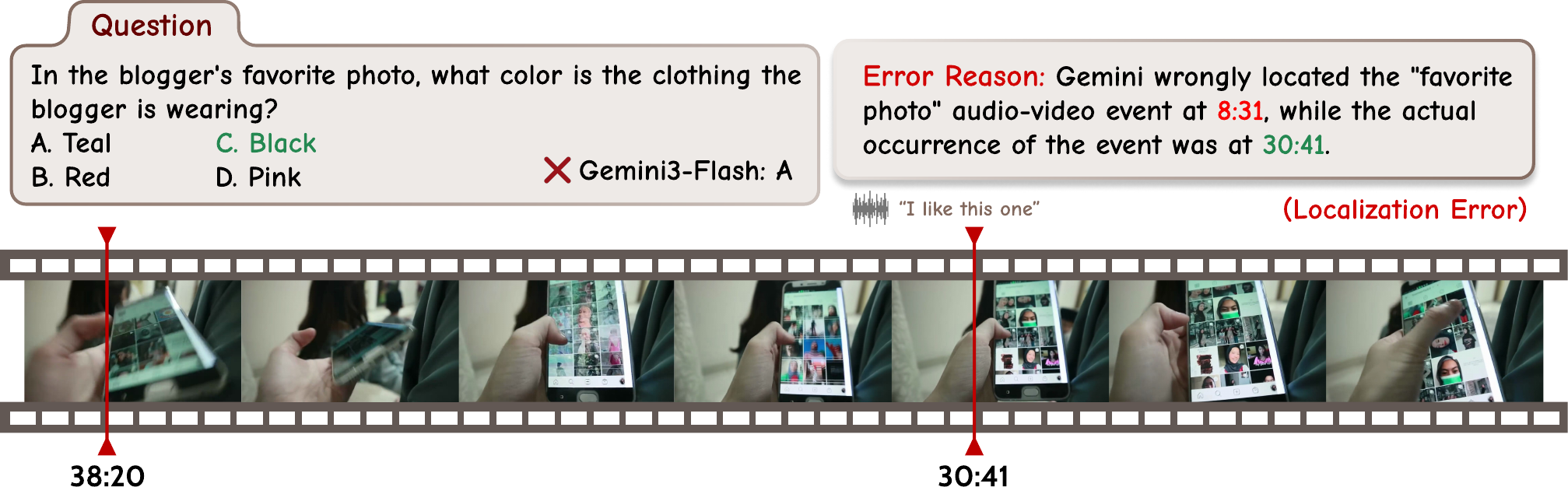}
    \caption{Cases of localization errors. 
    It is observed that models are unable to accurately locate the temporal occurrences of events within long-form audio and video inputs, leading to incorrect predictions.}
\label{fig:error_lo}
\vspace{-4mm}
\end{figure}

\begin{figure}[t]
    \centering
    \includegraphics[width=\linewidth]{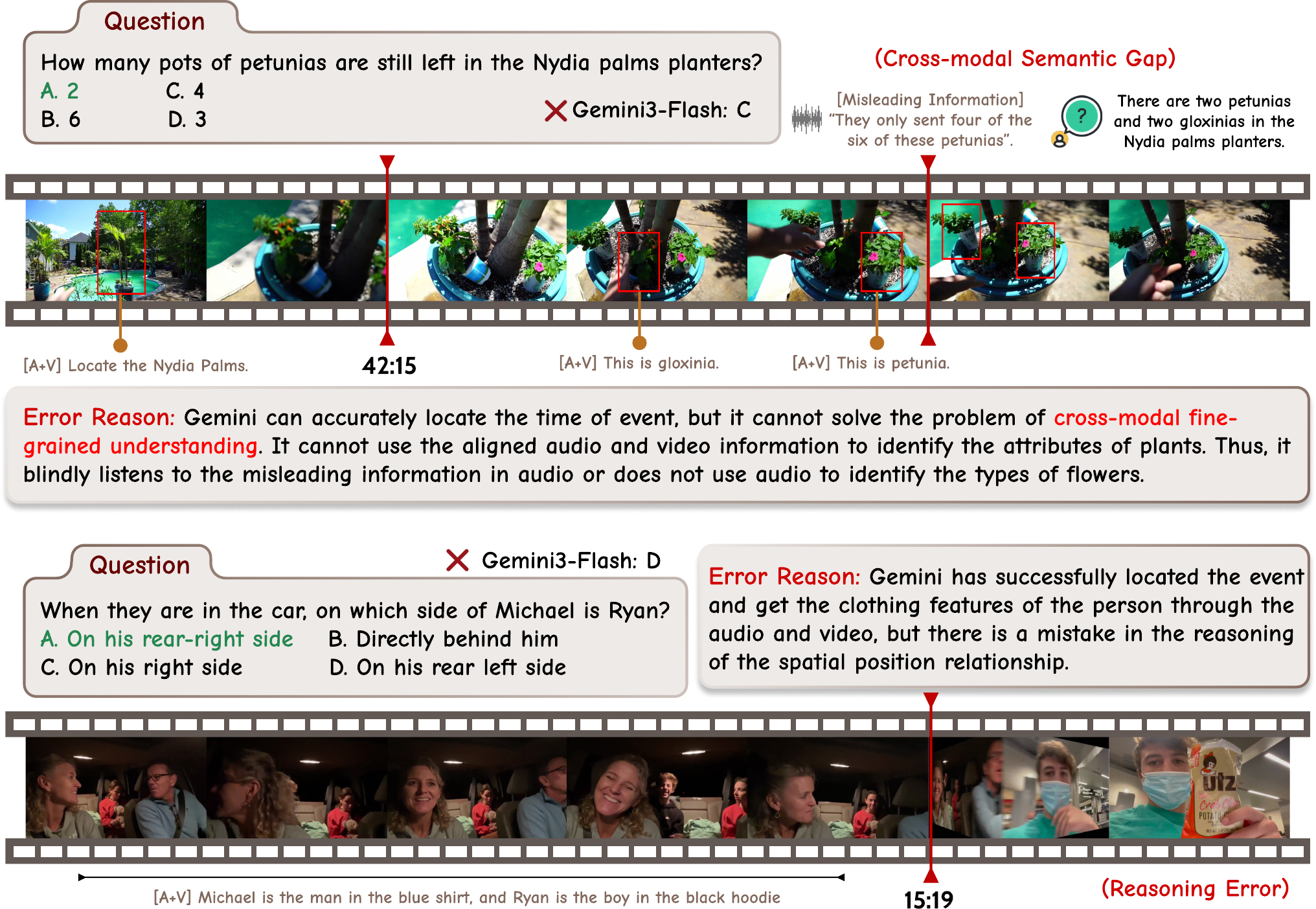}
    \caption{Cases of cross-modal semantic gaps and reasoning errors. 
    For complex problems, models fail to fully exploit audio and video information; instead, they often over-rely on a single modality during reasoning, hindering the alignment of cross-modal features. 
    Addressing this limitation represents a critical direction for future enhancements to OmniLLMs. 
    Furthermore, models frequently exhibit errors in reasoning tasks, such as spatial localization.}
\label{fig:error_cro}
\vspace{-4mm}
\end{figure}

In this section, we present representative cases across various error types to illustrate the challenges models face in contemporary audio-visual understanding.
Specifically, \cref{fig:error_per} depicts perception errors across both audio and visual modalities, while \cref{fig:error_lo} illustrates localization errors.
Furthermore, \cref{fig:error_cro} demonstrates the cross-modal semantic gap and highlights reasoning errors.
Overall, enhancing the capabilities of OmniLLMs necessitates improvements in long-form audio-visual comprehension, ranging from basic perception and localization to advanced reasoning and cross-modal understanding.
Consequently, substantial room for improvement remains, and LVOmniBench aims to provide valuable insights and guidance for future research.

\section{More Insights and Future Work}

\noindent \textbf{The Efficiency of Long-Context Capabilities.}
In the realm of long-video understanding, excessively long contextual token sequences pose a significant challenge, as they complicate model inference and incur substantial computational overhead.
While numerous established strategies have effectively addressed the long-sequence bottleneck for the video modality~\cite{chen2024image,liu2025global,tao2025dycoke,shen2024longvu, token_compression_survey,shao2025holitom,tao2025omnizip,yin2025videoarm,ding2025videozoomer,timechatonline,shang2024llava,liu2025video,ding2026omnisift,chen2025streamingtom,jin2025mergemix}, this issue persists for the audio modality.
Specifically, audio presents a more formidable compression challenge due to the strict temporal continuity of its signals.
Consequently, the development of efficient processing mechanisms for joint long audio-video inputs remains a critical open problem.

\vspace{1mm}
\noindent \textbf{Time Grounding Capabilities and Benchmarks.}
Our experiments demonstrate that the ability of models to perform temporal localization in long-form audio and video requires significant improvement, as a majority of errors stem from event localization failures during initial inference.
In the context of audiovisual understanding, audio event localization and joint temporal alignment across the two modalities are required, in addition to purely visual scene localization~\cite{wang2025time,wu2025survey,zhang2025timelens,zeng2024timesuite,qu2024chatvtg}.
However, research addressing temporal localization across omnimodal remains scarce~\cite{chen2025chronusomni}.
Consequently, the lack of comprehensive test benchmarks and methodological strategies constitutes a significant gap in current research.

\vspace{1mm}
\noindent \textbf{Cross-Modal Alignment and Fine-Grained Understanding.}
Furthermore, we observe that the ability of current models to integrate visual and audio information requires significant improvement.
In certain instances, models exhibit an over-reliance on information from a specific modality.
More specifically, OmniLLMs frequently ignore information from one modality by disproportionately prioritizing the features extracted from the other.
Consequently, this modality bias introduces substantial errors during cross-modal feature alignment.
Additionally, the alignment training of audio and visual modalities remains highly complex, which makes the achievement of fine-grained cross-modal comprehension a persistent challenge.

\vspace{1mm}
\noindent \textbf{Abstract Audio Perception Capabilities.}
Our experiments further reveal that models continue to exhibit significant errors in fine-grained perception of audio attributes, including frequency of occurrence, timbre, emotional tone, and sound pressure levels.
Furthermore, such non-linguistic information cannot be transcribed into text for comprehension via ASR or alternative methods.
Consequently, we emphasize the importance of future efforts dedicated to enhancing the ability of models to comprehend raw audio signals.

\section{License}
LVOmniBench is released under the CC BY-NC-SA 4.0 License. 

\section{Evaluation Prompt}

For the normal evaluation of open-source models and proprietary models, we follow the following prompt:
\begin{promptbox}
Question: $\langle$question$\rangle$ \\
Options: $\langle$options\_str$\rangle$

Select the best answer from the options above. Directly provide the letter representing your choice (A/B/C/D) and nothing else. Do not include the full text of the option; do not provide any explanation.
\end{promptbox}